\documentclass[a4paper]{article}
\usepackage{INTERSPEECH2021}
\usepackage{amsfonts}
\usepackage{amsbsy}
\usepackage{amsmath}
\usepackage{microtype}
\usepackage[ruled,vlined]{algorithm2e}
\usepackage{xcolor}
\usepackage{makecell}
\usepackage{multirow}
\usepackage{adjustbox}
\usepackage{rotating}
\usepackage{multirow}
\usepackage{hhline}
\usepackage{graphicx}
\usepackage{hyperref}
\usepackage{tipa}

\title{Grapheme-to-Phoneme Transformer Model for Transfer Learning Dialects}
\name{Eric Engelhart, Mahsa Elyasi, Gaurav Bharaj}
\address{
  AI Foundation, USA
}
\email{{\{ericengelhart, mahsa, gaurav}\}@aifoundation.com}

\begin{document}
\maketitle

\begin{abstract}
Grapheme-to-Phoneme (G2P) models convert words to their phonetic pronunciations. Classic G2P methods include rule-based systems and pronunciation dictionaries, while modern G2P systems incorporate learning, such as, LSTM and Transformer-based attention models. Usually, dictionary-based methods require significant manual effort to build, and have limited adaptivity on unseen words. And transformer-based models require significant training data, and do not generalize well, especially for dialects with limited data.

We propose a novel use of transformer-based attention model that can adapt to unseen dialects of English language, while using a small dictionary. We show that our method has potential applications for accent transfer for text-to-speech, and for building robust G2P models for dialects with limited pronunciation dictionary size.

We experiment with two English dialects: Indian and British. A model trained from scratch using 1000 words from British English dictionary, with 14211 words held out, leads to phoneme error rate (PER) of 26.877\%, on a test set generated using the full dictionary. The same model pretrained on CMUDict American English dictionary~\cite{weide2005carnegie}, and fine-tuned on the same dataset leads to PER of 2.469\% on the test set.
\end{abstract}


\section{Introduction}
\label{sec:intro}
Grapheme-to-Phoneme (G2P) systems convert sentences to their phonetic representation, and are an integral part of speech systems, like Text-To-Speech (TTS). While earlier G2P systems were rule-based \cite{elovitz1976automatic}, we now employ LSTM-based encoder-decoder models\cite{rao2015grapheme}, LSTM models with attention\cite{peters2017massively}, and finally transformer-based self-attention models\cite{yolchuyeva2020transformer}. G2P systems support multiple languages, and can be trained on several languages to increase data availability, for low-resource languages~\cite{vesik2020model,peters2017massively}. However, existing systems still have limitations.  

Dictionary-based approaches, with neural-network based fallback, have the advantage of reliability in pronunciations for words in the dictionary. While neural networks lead to higher accuracy at predicting words that are not in the dictionary, this approach fails when the dictionary is not large enough. The problem two-fold: 1) many words are not in the dictionary, and 2) the neural-network fallback is not robust with small training datasets. The following scenarios are difficult for such systems to handle:
\begin{itemize}
\item New words that do not match existing pronunciation, e.g. ``meme''.
\item Long words that neural networks trained on a single word fail to generalize to, e.g. ``supercalifragilisticexpialidocious''.
\item Spelling mistakes, or alternate spellings of the same word not seen in the dictionary, e.g. ``lite'', ``kool''.
\end{itemize}

Modern approaches use full sequence-to-sequence models to predict pronunciation of the entire inputs, and not just individual words. These approaches can be more robust to misspellings and leverage larger context for better pronunciation prediction. However, these approaches require larger datasets than dictionary-based approaches; as larger, more powerful, neural-network models require significantly more data to generalize.

In our work, we focus on dialects of English for G2P. This is an aspect of G2P that remains relatively less explored, but important for applications like TTS systems, where speakers who talk in different dialect than the G2P system could potentially have less optimal results. The speech synthesis model in this scenario is forced to learn the mapping between the dialect used in the G2P system, and the dialect the speaker actually speaks. This leads to model capacity inefficiencies on the mapping, that could have been used to better model a specific speaker's idiosyncrasies.

In this work, we show that limitations of modern transformer-based G2P systems can be overcome for English dialects that have limited dictionary data availability by transfer learning G2P models across dialects. We show how to generate training data for our system, and test our system using publicly available data. We examine the advantages of our proposed method, as the size of the target dictionary varies.

\section{Related Works}
While older G2P systems were rule-based~\cite{elovitz1976automatic}, joint-sequence models \cite{BISANI2008434} became the new normal, as they do not require expert labor to specify hundreds of rules. Neural sequence-to-sequence methods when applied to G2P take the form of LSTM-based encoder-decoder models \cite{rao2015grapheme}. While they were initially used to predict pronunciation of individual words, as neural machine translation evolved, newer methods were applied to G2P.

LSTM-based encoder-decoder models that augment attention between encoder and decoder, allow the models to convert whole sequences at once, instead of dealing with single words \cite{peters2017massively}. As transformer-based architecture became popular \cite{Vaswani2017AttentionIA}, they were also applied to G2P \cite{yolchuyeva2020transformer}; particularly in multilingual domain \cite{yu2020multilingual,vesik2020model}. Training a transformer-based model for dialectal G2P has been shown in \cite{blandon2019multilingual}, however, they train a model for a single dialect using a multilingual G2P, and do not investigate transfer-learning, or multiple dialects.

For low-resource languages, the dataset that a single language provides did not suffice the large data requirements of transformer-based architectures, so such systems are trained on multiple languages, to help alleviate the data requirements. Our focus, rather than on the multiple language case, lies at handling dialects of a single language -- English. Here, several dialects have extremely limited datasets, and we study how the transformer model behave on cross-dialect transfer learning (in English), as the size of the training dictionary varies. To the best of our knowledge, the problem of single language cross-dialect G2P transfer has not been studied in earlier works.
\begin{figure}[hbt!]
  \includegraphics[width=\linewidth]{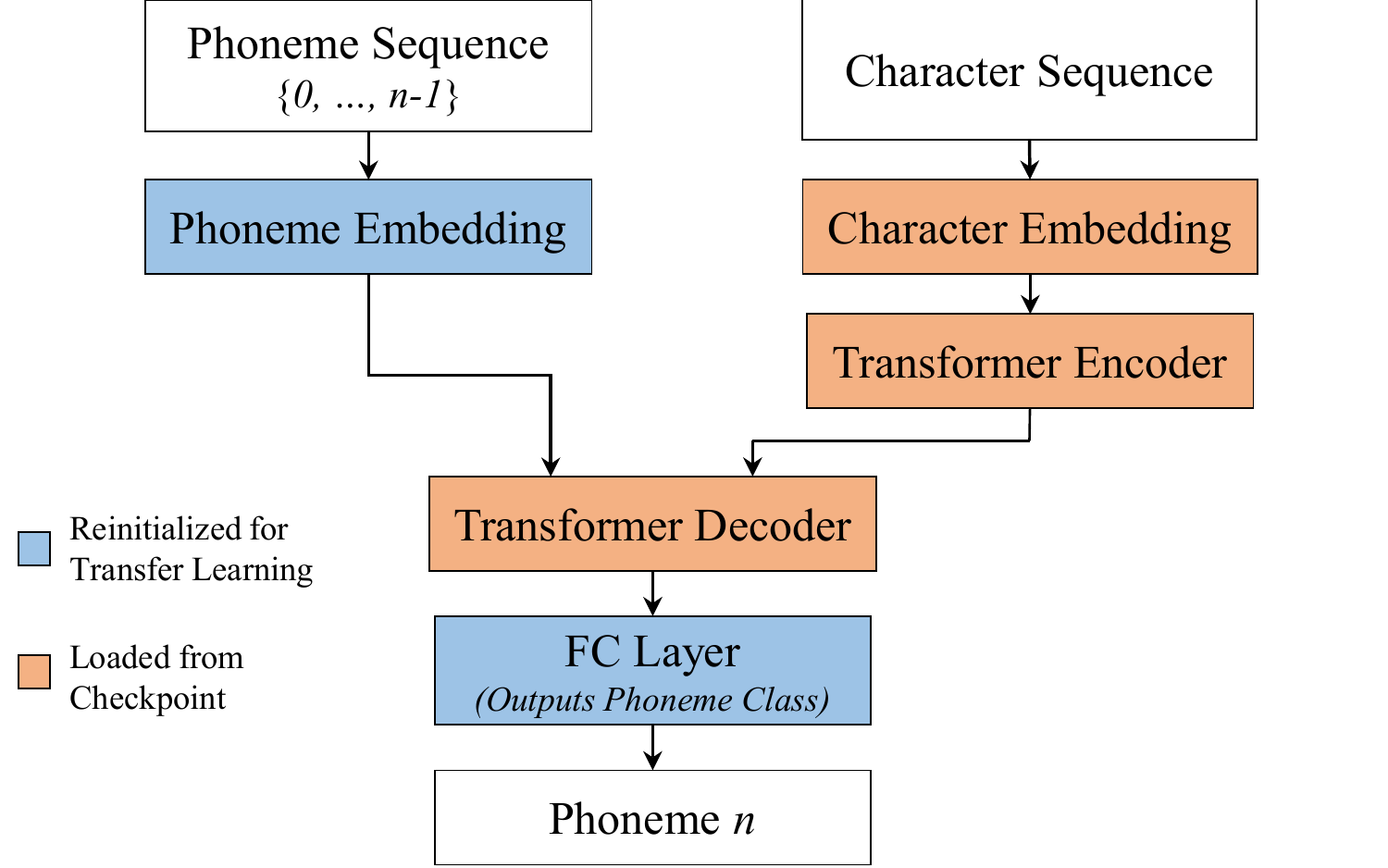}
   \caption{Diagram of the transformer model used, highlighting the layers that were replaced for transfer-learning and the layers that were kept the same from the pretrained model checkpoint.}
   \label{fig:g2pmodel}
\end{figure}
\begin{table}[h!]
\caption{Transformer Model hyper-parameters}
 \label{tab:table1}
  \begin{center}
\begin{tabular}{rc} 
\toprule
\textbf{Hyper Parameter} & \textbf{Value}  \\ 
\toprule
Encoder Layers                               & 3               \\
Decoder Layers                               & 3               \\
Hidden Width                                 & 256             \\
Position-wise Feedforward Width              & 512             \\
Attention Heads                              & 8               \\
Dropout Probability                          & 0.1             \\
Sequence Length                              & 240             \\
Normalization                                & Layer Norm      \\
\toprule
\end{tabular}
  \end{center}
\end{table}
\section{Model}
The Transformer architecture is a successful generic sequence-to-sequence model designed for neural machine translation. As G2P can be seen as a translation problem, translating graphemes to their phonetic counterparts, the Transformer architecture can be applied to G2P problems ~\cite{yolchuyeva2020transformer,vesik2020model,vesik2020model}. We use the Transformer architecture, ~\cite{Vaswani2017AttentionIA}, for our experiments, also see Figure~\ref{fig:g2pmodel}. We use the hyper-parameters specified in Table~\ref{tab:table1} for all experiments. During training, the hyper parameters are fixed across all models as seen in Table ~\ref{tab:table2}. The resulting Transformer model has $4.1$M parameters.
We train our Transformer models in the following setup(s):
\begin{enumerate}
    \item Pre-train the Transformer model using our CMUDict American English dialect data for 100 epochs, and save the model with the best validation loss.
    \item Train the Transformer model from scratch on both dialects, varying the dictionary size across [1K, 3K, 5K, 10K] words. We train the model for 50 epochs and select the model with the best validation loss for testing.
    \item Load the pre-trained Transformer model from setup 1. Then replace the embedding and output FC layers in the Decoder with new layers of the appropriate size for the differing output phoneme sets (see Figure~\ref{fig:g2pmodel} for details), dependent on dialect.  Finally, we fine-tune this model for 25 epochs on the same dialect and training dictionary size combinations as in setup two, again selecting the model with the best validation loss at the end of training.
\end{enumerate}

\begin{table}[h!]
 \caption{Training hyper-parameters}
 \label{tab:table2}
  \begin{center}
\begin{tabular}{rc} 
\toprule
\textbf{Hyper Parameter} & \textbf{Value}  \\ 
\toprule
Learning Rate            & 0.0005          \\
Batch Size               & 128             \\
Gradient Norm Clipping   & 1.0             \\
Initialization           & Xavier Uniform  \\
Optimizer                & Adam            \\
Random Seed              & 1234            \\
\toprule
\end{tabular}
  \end{center}
\end{table}
\section{Experiments}
\begin{figure*}
  \includegraphics[width=\linewidth]{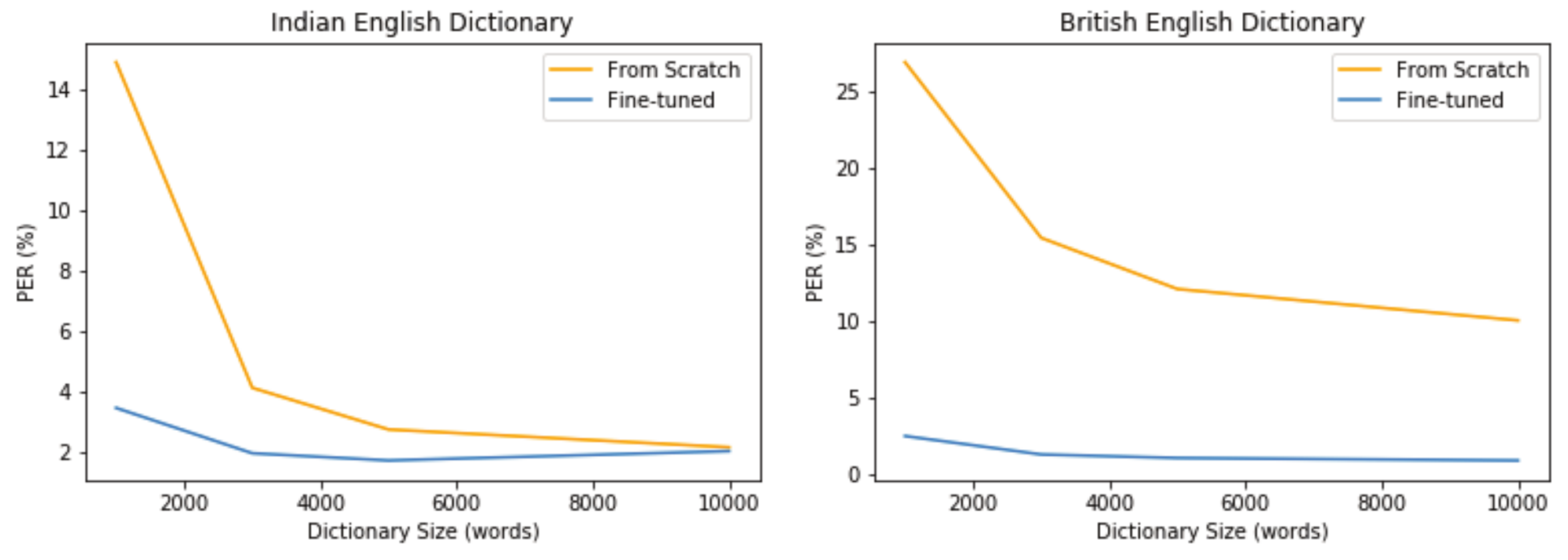}
   \caption{Comparing training from scratch and transfer learning as dictionary size increases.  (Left) Results for the Indian English dictionary.  (Right) Results for the British English dictionary.}
   \label{fig:britishresults}
\end{figure*}

\subsection{Dataset Creation} 
We use publicly available dictionaries to create our training and test datasets. We also avoid data automatically scraped from websites like Wikitionary\footnote{\url{https://en.wiktionary.org/wiki/Wiktionary}}, as such data is \emph{noisier} and less reliable~\cite{schlippe2010wiktionary}. We use CMUDict\cite{weide2005carnegie} and Indian English dictionaries from the CMUSphinx project. We also use Britphone dictionary by Jose Llarena\footnote{\url{https://github.com/JoseLlarena/Britfone}} for our British English (Received Pronunciation) dictionary. For every word with more than one pronunciation, that is, specifically not marked for a certain part of speech (in CMUDict's homonyms list), we take the first appearance of the word as its pronunciation. We tackle the G2P problem with full utterances to their phonetic transcriptions as a sequence-to-sequence problem, not as individual words but as entire sentences. To generate our training data, we take sentences from the LibriTTS\cite{zen2019libritts} dataset and split sentences for training/dev/test sets accordingly. We use the phoneme dictionaries to generate ground-truth phonetic sequences that correspond to the original sentences.

To make our training/test distributions more similar to real world scenarios, where the dictionary used for training has limited words compared to all sentences, we limit the number of words the dictionary has for training data, and filter out sentences that have any words not in this limited dictionary data. For testing our models, all words in the dictionary are used to generate the data, filtering out sentences that have words not in the dictionary.

To ensure that even extremely small subsets of large dictionaries contain common words such as ``the'' or ``of'', we compute the frequency of every word in the training set, and add words to the dictionaries based on their frequency. This helps build realistic \emph{pseudo-small} dictionaries from larger dictionaries. This is done because when dictionary for new dialect is built, the most common words in that dialect retain higher priority and increase the sentence coverage of that dictionary.

The 3\% shortest and 3\% longest sentences are filtered out of the training, validation, and test sets. This is done because the shortest sentences are single word utterances that often have more punctuation than letters. The longest sentences are filtered out to accommodate the maximum sequence length of the transformer model used, as extremely long sequences are computationally and memory intensive.
\subsection{Transfer Learning for English Dialects.} 
We generate dictionaries of 4 sizes for training: 1K words, 3K words, 5K words, and 10K words.  As the British English dictionary has 15221 words not counting alternate pronunciations, we do not train with any larger vocabulary.  For each dialect, a model is trained from scratch on each dictionary size, and a model is fine-tuned on each dictionary size.  In total, we train 16 different models for comparison.

We pretrain a model on data generated by the American English (CMUDict) dictionary for 100 epochs, selecting the lowest validation loss model as our checkpoint. We initialize our \emph{transfer-learned} models with the pretrained checkpoint.  We replace the output layers to match the new phonetic output size, and the whole model is fine-tuned for 25 epochs on the target dialect data. The checkpoint with the best validation loss is selected for testing. We also train the same model from scratch for 50 epochs to compare the utility of transfer learning.  At the end of every epochs, the validation loss is calculated on the dev set, and the final checkpoint is chosen based on lowest validation loss.

The models we train overfit to the data with small dictionary sizes, both from scratch and transfer learned, requiring validation loss checkpoint selection.  However, as the dictionary size increases, the overfitting decreases, and the Indian English models for 10K are likely undertrained with the standard number of epochs used for all models. The transfer learned models are able to achieve significantly lower validation and training loss before starting to overfit, this is more true for smaller dictionary sizes and the British English data, where the dictionary was able to generate far fewer sentences than the Indian English dictionary was.  For example, with a dictionary size of 5K words, the British English dictionary generated almost 9K training sentence-phoneme pairs, while the Indian English dictionary generated 87K sentence-phoneme pairs.  This discrepancy in data scale is likely why the Indian English from scratch models start to attain performance competitive with pretrained models at 10K words, while the British English from scratch models are still an order of magnitude worse in PER compared to the fine-tuned models.

\subsection{Results}
Our results show that transfer learning is effective means for building robust G2P models on dialects with extremely limited data. 
\\
\\
\textbf{British English Dictionary} The British English dictionary results shows this most effectively, as the dictionary size is inflated due to differing spellings across American and British English, e.g., ``color'' vs ``colour''. Even with this inflation of words, the fine-tuned British English dictionary of 1K words results in a low PER of only 2.469\%, see Figure~\ref{fig:britishresults}, right. As the full British English dictionary only has 15211 words when counting unique words, the improvement of the fine-tuned model is significant, as unlike CMUDict and the Indian English dictionary, that are large and have vocabularies of over 100K words. We note that the British English dictionary itself is not sufficient to train a robust G2P model from scratch, since, even with a training dictionary of size 10K words, the \emph{from scratch} model reaches 10.017\% PER, and the fine-tuned model achieves 0.875\% PER. The British English model suffers from a lower amount of training data per dictionary size as compared to the Indian English data, likely due to the LibriTTS dataset containing sentences more similar to the American English dialect than the British English dialect.
\\
\\
\textbf{Indian English Dictionary.} The Indian English dictionary results show that dictionaries without multiple spellings of the same word can have the gap quickly narrowed as the training data rapidly increases with dictionary size. With a 10K word dictionary for Indian English, a significant amount of training data is generated (\textgreater 100K training sentence-phoneme pairs). Models trained on larger amounts of data may benefit from more training epochs than were, in our experiments. The Indian English model fine-tuned on the 10K word dictionary data with the best validation loss came from the last epoch of training, while more training may have improved its performance. This explains its anomalous increase in PER compared to the other results from the Indian and British dictionaries. Despite the likely under-training of the models with large dictionary sizes for Indian English, the fine-tuned model's results (2.039\% PER) still improves upon the PER of the from scratch model (2.166\% PER) with half the training steps, making it much more computationally efficient, Figure~\ref{fig:britishresults}, left.

\subsection{Edge Cases}

As described in the Section~\ref{sec:intro}, dictionary-based G2P systems can run into edge cases, that can lead to inaccuracies. The main cause are: new words with confusing pronunciation, long words that cause the fallback neural network to fail, and 3) words that are misspelled and alternate spellings of the same word. Such edge cases can cause G2P to fail, and subsequent TTS is incorrect.

An example of the various edge cases is the poem \emph{Jabberwocky by Lewis Carroll}\footnote{\url{https://www.poetryfoundation.org/poems/42916/jabberwocky}}, which is full of invented words that lack meaning. These words pronunciation are guessed by the readers, although the rhyming structure of the poem does provide some clues. Our pretrained, and transfer-learned models are more capable of pronouncing this poem than dictionary systems that fail on words not in their vocabulary. We also note that some of the ``from scratch'' models have trouble with this poem, and the dictionaries used to train these systems would fail with invented lexicon of this poem. For example, for the original sentence:

\emph{\textcolor{blue}{``Beware the Jubjub bird, and shun The frumious Bandersnatch!''}}
\\
\\
\textit{British English 10K words -- Fine-tuned model} gives:

\textcolor{blue}{\textipa{b I w " E @ D " i : d Z " 5 b b " 3 : d , @ n d S " 5 n D " i : f \*r " u : m I @ s b " \ae n d @ z n " \ae tS !}}
\\
The fine-tuned model makes mistakes, for example, the word ``jubjub'' is translated to pronunciation as just \textipa{d Z " 5 b}, or ``jub''; a failure to repeat the sound again as the original word calls for.
\\
\\
\textit{British English 10K words -- From scratch model} gives:

\textcolor{blue}{\textipa{b j " u : D " E @ d Z " u : t b " 3 : d , \*r " 5 n d D @ S " a U n b l " A : f I s m @ \*r " e I S @ n s t s!}}
\\
In this phonetic transcript, the only word that can be identified with unambiguously correct pronunciation is the word ``bird'', and is translated to \textipa{[b " 3 : d]}. In this case, rest of the phonetic pronunciation is unclear, and only vaguely resembles the words the model was given. We also compare the longest word between the two models, \emph{\textcolor{blue}{bandersnatch}}. As it is the last word, it takes the last 9 phonemes from the models:
\begin{enumerate}
    \item[-] Fine-tuned model: \textcolor{blue}{\textipa{b " \ae n d @ z n " \ae t S}}
    \item[-] From scratch model: \textcolor{blue}{\textipa{@ \*r " e I S @ n s t s}}
    \item[-] Ground Truth pronunciation\footnote{\url{https://www.collinsdictionary.com/dictionary/english/bandersnatch}}:  \textcolor{blue}{\textipa{'" b \ae n d @ "" s n \ae t S}} 
\end{enumerate}
\section{Conclusions}

We propose a transfer learning based approach to cross-learn dialects of English, using a neural G2P model with a Transformer architecture, and attain high accuracy for dialects with smaller available data. We evaluate the PER across two dialects of English, sub-setting the dictionaries in a realistic manner to create artificially small training datasets, allowing for evaluation across dataset size. Our experiments show that for small dictionaries, transfer learning from models pretrained on larger dialects is a powerful method to significantly improve PER and quality. We observe that as the dictionary size increases, training a model from scratch quickly improves. However, depending on the dictionary, and the amount of training sentence-phoneme pairs it can generate, fine-tuning a model can lead to high accuracy. In the future, we plan to investigate application of multilingual approaches to multiple dialects, and instead create multi-dialect G2P models for multi-dialect/multi-speaker TTS models.

\bibliographystyle{IEEEtran}
\bibliography{0.main.bib}
\end{document}